\documentclass{svproc}
\usepackage{url}
\usepackage{pgfplots}  
\usepackage{multirow} 
\usepackage{multicol}
\usepackage[T1]{fontenc}
\usepackage{amsfonts}
\usetikzlibrary{patterns}
\begin{document}
\title{Matching Exemplar as Next Sentence Prediction (MeNSP): Zero-shot Prompt Learning for Automatic Scoring in Science Education }

\titlerunning{Matching Exemplar as Next Sentence Prediction (MeNSP)}
% \author{Anonymous Author(s) and Institution(s)}
\author{Xuansheng Wu$^{1}$ \and
Xinyu He$^{2}$ \and
Tianming Liu$^{1}$ \and
Ninghao Liu$^{1}$ \and \\
Xiaoming Zhai$^{2}$\thanks{Corresponding Author: xiaoming.zhai@uga.edu, ninghao.liu@uga.edu, tliu@uga.edu}}

\authorrunning{Wu et al.}

% \institute{Anonymous Institution(s)}
\institute{$^{1}$School of Computing, University of Georgia, Athens, GA, USA \\
$^{2}$AI4STEM Education Center, University of Georgia, Athens, GA, USA \\
\email{\{xuansheng.wu, xinyu.he1, tliu, ninghao.liu, xiaoming.zhai\}@uga.edu} \\
% \email{xinyu.he1@uga.edu} \\
% \email{tliu@uga.edu}\\
% \email{ninghao.liu@uga.edu}\\
% \email{xiaoming.zhai@uga.edu}
}

\maketitle              % typeset the header of the contribution
\begin{abstract}
Developing natural language processing (NLP) models to automatically score students' written responses to science problems is critical for science education. However, collecting sufficient student responses and labeling them for training or fine-tuning NLP models is time and cost-consuming. Recent studies suggest that large-scale pre-trained language models (PLMs) can be adapted to downstream tasks without fine-tuning by using prompts. However, no research has employed such a prompt approach in science education. As students' written responses are presented with natural language, aligning the scoring procedure as the next sentence prediction task using prompts can skip the costly fine-tuning stage. In this study, we developed a zero-shot approach to automatically score student responses via \textbf{M}atching \textbf{e}xemplars as \textbf{N}ext \textbf{S}entence \textbf{P}rediction (MeNSP). This approach employs no training samples. We first apply MeNSP in scoring three assessment tasks of scientific argumentation and found machine-human scoring agreements, Cohen's Kappa ranges from 0.30 to 0.57, and F1 score ranges from 0.54 to 0.81. To improve scoring performance, we extend our research to the few-shots setting, either randomly selecting labeled student responses at each grading level or manually constructing responses to fine-tune the models. We find that one task's performance is improved with more samples, Cohen's Kappa from 0.30 to 0.38, and F1 score from 0.54 to 0.59; for the two other tasks, scoring performance is not improved. We also find that randomly selected few-shots perform better than the human expert-crafted approach. This study suggests that MeNSP can yield referable automatic scoring for student-written responses while significantly reducing the cost of model training. This method can benefit low-stakes classroom assessment practices in science education. Future research should further explore the applicability of the MeNSP in different types of assessment tasks in science education and further improve the model performance. Our code is available at \url{https://github.com/JacksonWuxs/MeNSP}.

\end{abstract}

%1. Assessment: 题目来评估学生的能力 （工具）
%2. written Response: 简答题 （形式）
%3. Argumentation: 我们现在评估的能力 (目标）
%4. Rubrics: 得分点
%5. Scoring: Student Response ----> Teacher Grades
%6. Examples: 由出题人给的示例response

\section{Introduction}
To engage students in meaningful science learning, it is crucial that science assessments can elicit student knowledge-in-use so that instructors can better understand and support student learning~\cite{Pellegrino2013-hr}. The need for assessing student competence in using scientific knowledge to solve problems or design solutions has been identified in the actualization of 21$^\mathrm{st}$-century science education~\cite{national2012framework}. While such science and engineering practices are desirable for students to enact, it has been challenging to assess with multiple-choice questions. More complicated written response assessments are needed to assess such scientific practices.

Although the written response assessment allows for the freedom of expression elicited in responses, it accounts for a number of challenges in the process of scoring. Since scoring is expected to be fair and reliable, it requires, among other efforts, that educators are trained using pre-designed rubrics to enhance reliability. Despite the expertise and rubric familiarity of the raters, it demands a lot of time to score the learners' responses appropriately.

In contrast to some of the issues associated with human scoring, machine learning, such as Natural Language Processing (NLP), offers quicker, less expensive, and more consistent scoring~\cite{zhai2021practices}. However, most NLP scoring models need to be tested with human scoring as benchmarks before applying in classroom settings. Their validity and performance are a function of the former~\cite{Wolfe2020-ul,zhai2021validity}. The majority of the models needed to achieve this require a large sample of data for training, validation, and testing purposes~\cite{zhai2020applying}. Given that the conditions necessary to collect a reasonable size of data are a huge effort in the field of education, satisfying the needs of these models is cumbersome~\cite{zhai2021meta}. 

To overcome these challenges, this study employs prompt learning with pre-trained language models (PLMs) to develop algorithmic models for automatic scoring. Prompt learning utilizes little or limited labeled data (i.e., zero-shot) in developing a model and has shown great potential in accomplishing NLP tasks with significant efficiency~\cite{zhong2021adapting}. To verify the efficiency of this new approach, we develop NLP scoring models to automatically score students' written arguments to science phenomena when engaging in scientific argumentation practices.

% Motivation1: It is difficult to collect large sample data in the education field.  ---> prompt learning to break the constraint of training samples.

% Motivation2: It is cost-consuming to develop a perfect educational dataset for model training. (To guarantee the coherence between teacher graders is time- and cost-consuming.) ---> one rubric given by the task designer is much more reliable.
% In situations where students are not familiar with the type of tasks, the chances of collecting a normally distributed data would be low, therefore affecting the reliability of the test. Wolfe (2020) argued that after identifying and evaluating the discrepancies between human and machine scores, it is possible to have the machine scores advantageous. 

% Challenge: As students are not familiar with the new type of tasks, it is hard to generate data with a normal distribution. ---> two-stage prompt learning by using two different pretraining tasks.

This study will contribute to the efficiency of developing innovative assessments in education broadly and science education specifically. Our approach will significantly save the time and cost of developing NLP scoring algorithmic models for automatic scoring, which is a bottleneck that has prevented the broad use of machine learning-based assessment practices in classrooms~\cite{zhai2020applying}. Implementing this approach could benefit millions of students using automatically scored written response assessments that have been developed~\cite{harris2019designing}.

\section{Related Work}
\subsection{Natural Language Processing for Automatic Scoring }
Natural language processing (NLP) is a field of computer science using computational techniques to learn, understand, and produce human language content. There has been a long time to apply NLP in language education, such as to correct students' writing errors, conduct semantic analysis, or assess language skills~\cite{litman2016natural}. With the improvement of using NLP to evaluate students' writing skills, research has begun to explore using NLP to evaluate the content of students' writing and study the quality of writing at the level of domain-specifics such as science learning. written response assessment in science education is one of the specific domains needing NLP for automatic scoring~\cite{zhai2020applying}. 

To help teachers better understand students' scientific thinking, researchers have explored using NLP technologies to score student-written responses a decade ago automatically. Haudek et al.~\cite{haudek2012they} used SPSS Text Analytics for Surveys to score students' biology understanding automatically. The program can extract key linguistic features of student writing to classify student responses according to scoring rubrics. Researchers~\cite{nehm2012transforming} at Carnegie Mellon University developed SIDE (current version named LightSIDE) that integrates various algorithmic functions. Nehm et al.~\cite{nehm2012transforming} employed his package to develop a portal EvoGrader to examine students' understanding of biology concepts. Educational Testing Service developed a C-rater~\cite{liu2014automated} for the automatic scoring of GRE essays and short written-response answers. Later, they incorporated and upgraded the tool to C-rater-ML~\cite{gerard2019guiding,lee2019automated} and employed it to automatically score students' scientific argumentation and explanations. While prior tools employed individual algorithms, researchers~\cite{maestrales2021using,uhl2021introductory} also developed tools that ensemble multiple algorithms to score student-written responses simultaneously. Most recently, researchers~\cite{riordan2020empirical,amerman-a} also employed BERT~\cite{devlin2018bert} for automatic scoring of student-written responses for scientific practices.

Among these developments, recent surveys~\cite{zhai2021meta,zhai2020applying} suggested that developing these models requires a large number of human-scored written responses with varying accuracy. However, collecting and validating the student responses and rigorous rubrics for scoring takes much effort. Since training human experts to use the scoring rubric to assign scores to student responses reliably is challenging, obtaining datasets to develop these models is costly.

\subsection{Prompt Learning}
Prompt learning~\cite{brown2020language,su2022transferability,vu2021spot,wu2023survey} leads to a new paradigm in NLP as it can achieve comparable performance to full-parameter fine-tuning with fewer training samples and parameters. ``Prompt'' typically is a short piece of text that include instructions for the task (zero-shot learning) or a few samples of the task (few-shots learning)~\cite{mayer2022prompt}. By selecting appropriate prompts, it is possible to directly predict the target label using the pre-trained language models. 
Compared with the previous ``pre-train and fine-tune'' paradigm, prompt learning has the advantage of reducing the training cost and being applicable to multiple tasks by changing the prompts~\cite{liu2021pre}.

The new paradigm brings its advantage as well as a new challenge, that is, how to find the most appropriate prompt. The most natural way is to create intuitive prompts manually~\cite{schick2020s,liu2021gpt}. However, some researchers~\cite{shin2020autoprompt,liu2021pre} found that the optimal prompt may not be readable by humans. Thus, many methods~\cite{shin2020autoprompt,schick2020exploiting,liu2021gpt,liu2021p} are proposed to learn a better prompt automatically. Besides learning prompts, providing and ordering a few additional answers in the prompts can also result in satisfied model performance~\cite{gao2020making,lu2021fantastically}. 

To the best of our knowledge, the application of prompt learning in the education field is still at the beginning. Hart-Davidson et al. (2021)~\cite{omizo2021detecting} applied prompt learning in the qualitative coding research task, which could be used to provide written feedback for student writing.
Zhang et al. (2022)~\cite{zhang2022automatic} applied prompt learning in the fine-tuning process to boost the performance for automatic scoring of short Math answers. How to give full play to The advantage of using prompt learning hasn't been fully discussed.

In summary, there is a lack of research exploring PLMs' zero-shot performance for automatic scoring in education, not even in science education. 
Therefore, this study applies prompt learning to automatically score student-written responses to science assessments. The approach can be applied to more tasks in education to save time and cost.
%The performance of the prompt learning paradigm is closely related to how to format the downstream task to the pre-trained task. 

\section{Approach}
This section proposes a method to develop the system for scoring student-written arguments without fully labeled training datasets. 
Inherently, student responses are presented with natural language, which is readable by PLMs. 
Since PLMs are built under some pre-trained tasks over natural language corpus, we can reformat the scoring task as one of the pre-trained tasks so that the textual responses of students can be graded without further fine-tuning of the models on labeled responses.
Reformatting a downstream task as a pre-trained task is also known as \textit{prompt learning}, which has been widely applied to handle NLP tasks under the zero-shot and few-shots settings where labeled data are unavailable or limited. 
However, letting PLMs understand (1) the meaning of the ``scoring'' procedure, and (2) the scoring rubrics of each assessment item are two non-trivial challenges to achieving the goal.
Technically, we need to find a pre-trained task to reformat the scoring procedure and combine student responses and rubrics as new inputs to feed for PLMs.
In the rest of this section, we present a two-stage pipeline to score student responses based on the rubrics of the assessments.

\begin{figure}
% \vspace{-0.5cm}
\centerline{\includegraphics[width=0.99\linewidth]{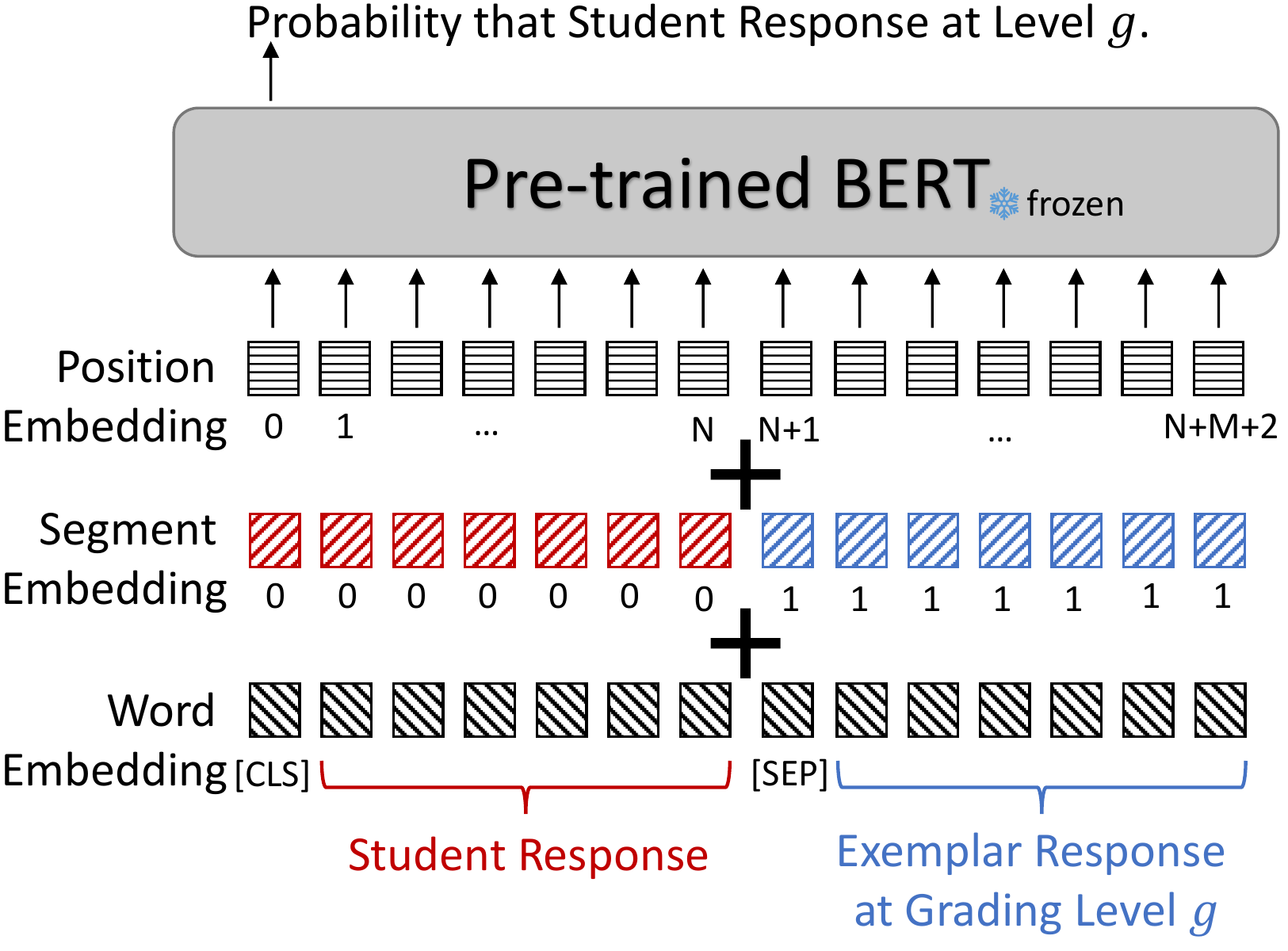}}
\caption{Next sentence prediction as matching exemplars.}
\label{main_figure}
\vspace{-1.cm}
\end{figure}

\subsection{Matching Exemplars}

Most researchers~\cite{shin2020autoprompt,liu2021gpt,liu2021p} reformat their downstream tasks as the masked language modeling task~\cite{devlin2018bert}, where PLMs are asked to fill the blank \emph{[MASK]} that is shown in a context with a proper word. 
For example, by combining a student response and the corresponding rubric, we can construct a new input as \emph{"The student responses are that \underline{[Response]}. The rubric is that \underline{[Rubric]}. Overall, the response can be grad as [MASK] point.''}, where each underlined word is filled with the actual input, and the rest parts of this example are called \textit{template}. 
Ideally, we expect the PLMs to fill the blank \emph{[MASK]} with the grade of the response according to the given scoring rubric.  
However, our piloting data show that PLMs almost randomly fill in the blank without considering the context.
We assert that this strategy is failed for two reasons. 
Firstly, both the responses and rubrics are so long that PLMs pay less attention to the task definition provided by the template. 
Secondly, PLM has no hints to bridge the gap between the rubrics and the final grading points. 

To some extent, scoring student responses according to a given rubric can be considered a matching process if the rubrics can be refined to each grading level. 
Following this idea, we propose reforming the scoring procedure as the Next Sentence Prediction (NSP) task~\cite{devlin2018bert}, which requires PLMs to determine whether the given two texts come from the same context. 
Specifically, assuming that the rubrics on each grading level are available, we first concatenate the student responses with each rubric's overall grading levels independently. 
After that, PLMs judge whether each response-rubric pair shows the same context by using the outputs of the NSP task. 
Finally, each score of a student response will be the grading level indicated by the rubrics that best match it. 
Since the objective of the NSP task has been clear to PLMs, we no more need to worry about designing templates to let the PLMs know what we require them to do. 
To this end, we address the first concern of using the MLM task: PLMs cannot easily identify what the scoring task is.
On the other hand, to fill the gap between the rubrics and the student responses, we propose replacing the rubrics with exemplars as standards to score the responses. 
This strategy is necessary because the language style of the student responses dramatically differs from that of the assessment rubrics, which leads PLMs to constantly predict the rubric-response pair as different (Here is a sample case\footnote{Rubric: Student can specify Sam's claim and corresponding evidence, and explain the relationship between the claim and evidence properly. Response: Sam's claim is that gas particles float on the top in the box. The evidence is that bubbles in soda water float to the top. As gas particles and bubbles, all refer to air. Sam can infer that air in the box can also float on the top just as air in soda water does.}). 
Figure~\ref{main_figure} demonstrates the proposed method.

Formally, given a pre-defined vocabulary set $\mathcal{V}$, a student response $\mathcal{R}=[w^{(1)}, ..., w^{(n)}, ..., w^{(N)}]$ consists of $N$ words $w^{(n)}\in\mathcal{V}$, and the $g$-level grade exemplar response $\mathcal{E}_g=[w_g^{(1)},...,w_g^{(m)},...,w_g^{(M)}]$ has $M$ words $w^{(m)}\in\mathcal{V}$. We adopt a PLM $f_\mathrm{nsp}: \mathcal{V}^L\rightarrow[0,1]$ with a pre-trained NSP head to automatically score the student response as:
\begin{equation}
 score(\mathcal{R}) = \arg\max_{g}\,f_\mathrm{nsp}([w_\mathrm{cls}; \mathcal{R}; w_\mathrm{sep}; \mathcal{E}_g])\,,
 \label{main_equ}
\end{equation}
where $w_\mathrm{cls}$,$w_\mathrm{sep}\in\mathcal{V}$ are special words, $[\cdot;\cdot]$ indicates the concatenating operation over the input words, and $L=M+N+2$ in this case. According to Equation~\ref{main_equ}, we achieve automatic scoring under the zero-shot scenarios (without training) by aligning the scoring process into an exemplar matching process and releasing the PLMs' ability learned from the pre-training stage.

\subsection{Zero Grade Identifier}
However, the above strategy raises a new challenge in obtaining fine-grained exemplars for each grading level.
We simultaneously cope with this challenge by decoupling the rubric and the perfect response of an assessment item. 
Particularly, the perfect rubric can be separated into several points, where each point reflects a grading level. 
Therefore, it is easy to write down the fine-grained rubrics of each grading level. 
Similarly, we first develop an optimal student response that fits all grading points of the perfect (full score) rubric.
Once the optimal response is given, we remove each part of the perfect response to generate exemplar responses at different levels gradually. 

Although this strategy generates high-quality responses for non-zero grading levels, it cannot and is also impossible to enumerate all zero-point exemplars since the reasons that the responses receive high scores are limited (exhaustive). In contrast, the zero-point responses can be various (in-exhaustive).
To reduce invalid scoring because of missing zero-point exemplars, we introduce a pre-stage before the above method so that we find out those zero-point responses in an early stage and let Equation~\ref{main_equ} focus on how many points the responses reward. 
Recall that PLMs represent inputs as vectors in their interior, carrying rich semantic information. 
Since the zero-point response is very different from the perfect response, the distance between the vectors of them should be far away. 
Thus, PLMs generate vectors to represent both the perfect exemplar response and the given student response and measure the distance between these two vectors with the Cosine similarity. 
If the cosine similarity between them is smaller than a threshold, we directly grade them with zero point. 
The threshold is calculated by averaging the cosine similarities of the zero-score and the one-score exemplar response to the perfect exemplar response.

Theoretically, given a PLM $f_\mathrm{emb}: \mathcal{V}^L\rightarrow \mathbb{R}^d$ that maps a piece of $L$-length text into a $d$-dimensional space, we collect the representations of the student response and the $g$-level exemplar $\mathbf{z}_\mathrm{R}=f_\mathrm{emb}(\mathcal{R})$ and $\mathbf{z}_g=f_\mathrm{emb}(\mathcal{E}_g)$, respectively. 
We determine whether the response $\mathcal{R}$ belongs to $0$-level grade by:
\begin{equation}
zero\_score(\mathcal{R})=\left\{
\begin{array}{rl}  
\textit{Yes}\,,&cos(\mathbf{z}_\mathrm{R}, \mathbf{z}_{g=2}) < \theta\,, \\
\textit{No}\,, & \textit{otherwise}\,,
\label{filter}
\end{array}
\right.
\end{equation}
where
\begin{equation}
\label{theta}
\theta = \frac{\cos(\mathbf{z}_{g=0}, \mathbf{z}_{g=2}) + \cos(\mathbf{z}_{g=1}, \mathbf{z}_{g=2})}{2}\,,
\end{equation}
$\cos(\mathbf{x},\mathbf{y})=\frac{\mathbf{x}\mathbf{y}^\mathsf{T}}{||\mathbf{x}||_2||\mathbf{y}||_2}$ is the cosine distance among vectors, and $\mathbf{z}_{g=i}\in\mathbb{R}^d$ is the embedding of the $i$-th grading level exemplar, . 
The combination of Equation~\ref{filter} and~\ref{theta} identifies most zero-score responses without enumerating all possible zero-level exemplars. Since Equation~\ref{filter} and~\ref{theta} are non-parametric, we can identify zero-score responses under the zero-shot setting.

\section{Experiment}
\vspace{-0.3cm}
This section aims to quantitatively justify the following three research questions (\textbf{RQ}): (1) How accurate is the proposed method in scoring student responses? (2) Can the performance of the proposed method be further improved if a few sample responses are available? (3) How does the quality of sample responses affect the method performance under the few-shots setting? 

\vspace{-0.1cm}
\subsection{Setup}
\vspace{-0.2cm}
\subsubsection{Dataset}
We choose a subset of an existing dataset of argumentation items~\cite{zhai2022assessing} for our experiments. 
The dataset originally consists of eight written response items sharing the same context regarding gases.
The items require varied levels of cognitive demands aligned with a learning progression of argumentation in science~\cite{osborne2016development}. 
Students' responses were scored based on their performance on claim stating, evidence clarifying, and warrants using. 
According to the different cognitive demands of each item, the scoring rubrics varied between 2 to 4 levels.
Meanwhile, each item is designed with different levels of complexity, diversity (refers to the three-dimensional science learning requirement), and structure (refers to the learning progression of scientific argumentation~\cite{haudek2021exploring}. 
In this study, we select three items (G4, G5, and G6) sharing the same diversity, structures, and rubrics level but distributing in two complexity levels as our downstream tasks.
Overall, the dataset contains 2081 labeled responses (770 for item G4, 669 for item G5, and 642 for item G6) from 931 students of grades 5 to 8. 

\paragraph{Dataset Splitting}
To mimic a few-shot setting, we only leave three samples at each grading level to answer \textbf{RQ2} and \textbf{RQ3}. 
Here, no valid set remains because tuning the hyper-parameters for each item is not encouraged. 
Since almost the entire dataset is considered the test set, the K-folds setting is unnecessary. 

\subsubsection{Metric}
We calculate the Cohen's kappa (Kappa) and F1 scores to measure the performance of the machine learning models for auto-scoring.
Cohen's kappa is one of the standard rater agreement indices to quantify levels of agreement between computer scoring and human expert scoring \cite{bejar1991methodology}. 
Kappa value ranges from -1 to 1, where 1 indicates a strong agreement between the human score and the machine score and 0 refers to an opposite agreement. 
Typically, a machine learning model can be accepted if it reaches around 0.4 Kappa score~\cite{zhai2020applying}.
%In this study, we examine the model's Kappa in two ways: 1) whether the kappa exceeds .40; 2) whether the kappa increases with increasing training samples. 
We also report the F1 score, the weighted harmonic mean of Recall and Precision, to evaluate the model performance \cite{powers2015f}. 

\subsubsection{Baseline}
We compare MeNSP with some baseline methods to measure the effectiveness of MeNSP.
Under the zero-shot setting, we perform \textbf{Random} strategy as our baseline, which randomly scores the student responses from 0 to 2 (three grading levels). 
Under the few-shots setting, we follow the previous studies~\cite{zhai2022assessing} that performed machine learning in the automatic scoring task. Specifically, we choose some popular ensemble models as the baselines, including Gradient-Gased Decision Tree (\textbf{GBDT}), Random Forest with Decision Tree (\textbf{RFDT}), and the simple Voting strategy (\textbf{Vote}) over five basic models (e.g., Naive Bayes, Decision Tree, Logistic Regression, Multilayers Perceptron, and Support Vector Machine). 
All baselines make decisions based on the TF-IDF~\cite{salton1988term} scores of words presented in the student responses.

\subsubsection{Exemplar Design}
We first manually develop exemplars for each task according to the rubric of the highest level (level 2)~\cite{haudek2021exploring}. 
Human experts are involved to ensure that the level 2 exemplars contain all elements of a perfect argument to the greatest extent. 
Then, we delete elements level-by-level and adapt sentences to meet the rubrics of level 1 and level 0. 
These exemplars are the prompts used in the zero-shot experiment. 

\subsubsection{Sample Design}
To examine the model performance training with few samples, we use two strategies to generate samples for the few-shots tuning: 
(1) We randomly select student responses for each level from the dataset. 
(2) We use ChatGPT\footnote{ChatGPT is available at https://chat.openai.com/chat.} to generate new responses based on the exemplar at each grading level and then conduct manual inspection and adjustment to ensure the machine-generated responses meet the rubrics.

\begin{table*}[]
  \caption{Machine automatically scoring performance.}
  \label{main_table}
  \centering
\begin{tabular}{c|c|c|cc|cc|cc}
\hline\hline
\multirow{2}{*}{\textbf{Shot}} & \multirow{2}{*}{\textbf{Sample}}  & \multirow{2}{*}{\textbf{Model}} & \multicolumn{2}{c|}{\textbf{G4}} &\multicolumn{2}{c|}{\textbf{G5}} &\multicolumn{2}{c}{\textbf{G6}} \\ 
&  &       & Kappa\,(\%) & F1\,(\%) & Kappa\,(\%) & F1\,(\%) & Kappa\,(\%) & F1\,(\%)   \\
\hline
\hline
\multirow{2}{*}{0} & \multirow{2}{*}{-} & Random & $-0.2_{\pm 3.4}$ & $32.9_{\pm 1.9}$ & $-1.1_{\pm 2.2}$ & $35.8_{\pm 1.5}$ & $-3.3_{\pm 3.6}$ & $30.6_{\pm 2.0}$ \\
                   &                    & MeNSP  &$30.3_{\pm 0.3}$ & $54.2_{\pm 0.2}$ & $57.2_{\pm 1.1}$ & $81.1_{\pm 0.5}$ & $34.5_{\pm 0.3}$ & $57.0_{\pm 0.3}$ \\

\hline
\multirow{8}{*}{1} & \multirow{4}{*}{Random} & RFDT & $-1.1_{\pm 2.9}$ & $23.8_{\pm 5.5}$ & $6.5_{\pm 9.4}$ & $50.0_{\pm 14.8}$ & $-1.1_{\pm4.5}$ & $24.1_{\pm 11.3}$ \\
                   &                         & GBDT  & $5.6_{\pm 7.2}$ & $34.7_{\pm 7.3}$ & $6.1_{\pm 16.2}$ & $43.6_{\pm 17.9}$ & $3.3_{\pm 9.9}$ & $34.6_{\pm 11.7}$ \\
                   &                         & Vote & $3.1_{\pm 11.2}$ & $33.6_{\pm 9.1}$ & $20.9_{\pm 12.2}$ & $62.0_{\pm 8.9}$ & $7.3_{\pm 9.5}$ & $36.5_{\pm 14.1}$ \\
                   &                         & MeNSP & $35.9_{\pm 3.2}$ & $57.9_{\pm 2.2}$ & $43.0_{\pm 10.1}$ & $74.3_{\pm 6.1}$ & $25.7_{\pm 4.3}$ & $53.2_{\pm 3.6}$ \\
                   \cline{2-9}
                   & \multirow{4}{*}{Manual} & RFDT & $0.1_{\pm 0.1}$ & $26.3_{\pm 0.1}$ & $0.3_{\pm 0.3}$ & $51.5_{\pm 0.1}$ & $0.0_{\pm 0.0}$ & $34.2_{\pm 0.0}$ \\
                   &                         & GBDT  &  $0.0_{\pm 0.0}$ & $26.2_{\pm 0.0}$ & $1.5_{\pm 1.5}$ & $52.5_{\pm 1.0}$ & $0.7_{\pm 0.3}$ & $34.8_{\pm 0.3}$ \\
                   &                         & Vote & $7.0_{\pm 1.7}$ & $34.0_{\pm 1.4}$ & $5.6_{\pm 0.7}$ & $52.2_{\pm 0.4}$ & $-0.14_{\pm 0.0}$ & $34.2_{\pm0.0}$ \\
                   &                         & MeNSP  & $35.5_{\pm 7.1}$ & $59.1_{\pm 1.8}$ & $27.0_{\pm 4.5}$ & $56.5_{\pm 0.5}$ & $30.9_{\pm 1.8}$ & $56.8_{\pm 0.6}$ \\
                        
\hline
\multirow{8}{*}{3} & \multirow{4}{*}{Random} & RFDT & $5.5_{\pm 6.2}$ & $33.3_{\pm 5.9}$ & $13.2_{\pm 14.4}$ & $49.8_{\pm 19.8}$ & $3.6_{\pm 2.2}$ & $33.1_{\pm 11.2}$ \\
                   &                         & GBDT  &  $4.0_{\pm 7.2}$ & $33.2_{\pm 7.4}$ & $14.0_{\pm 16.0}$ & $49.6_{\pm 19.9}$ & $13.6_{\pm 3.6}$ & $46.2_{\pm 1.7}$ \\
                   &                         & Vote & $15.6_{\pm 3.4}$ & $44.8_{\pm 2.0}$ & $20.9_{\pm 16.7}$ & $57.4_{\pm 10.8}$ & $13.1_{\pm 4.9}$ & $44.1_{\pm 6.9}$ \\
                   &                         & MeNSP  &  $37.5_{\pm 3.3}$ & $59.0_{\pm 2.0}$ & $51.6_{\pm 8.2}$ & $78.7_{\pm 3.6}$ & $27.8_{\pm 6.7}$ & $53.4_{\pm 4.6}$ \\
                   \cline{2-9}
                   & \multirow{4}{*}{Manual} & RFDT & $8.1_{\pm 3.9}$ & $34.2_{\pm 3.7}$ & $0.7_{\pm 0.4}$ & $51.6_{\pm 0.1}$ & $0.0_{\pm 0.0}$ & $34.2_{\pm 0.0}$ \\
                   &                         & GBDT  &  $17.1_{\pm 0.4}$ & $45.3_{\pm 0.3}$ & $7.6_{\pm 0.2}$ & $53.2_{\pm 0.1}$ & $1.03_{\pm 0.0}$ & $35.64_{\pm 0.0}$ \\
                   &                         & Vote & $14.8_{\pm 1.0}$ & $43.9_{\pm 1.0}$ & $8.1_{\pm 2.1}$ & $52.5_{\pm 1.1}$ & $7.83_{\pm 0.6}$ & $40.07_{\pm 0.4}$ \\
                   &                         & MeNSP  &  $33.8_{\pm 1.2}$ & $56.2_{\pm 0.9}$ & $24.8_{\pm 1.1}$ & $58.2_{\pm 1.4}$ & $30.1_{\pm 0.5}$ & $56.9_{\pm 0.3}$ \\
\hline
\hline
\end{tabular}
\vspace{-0.6cm}
\end{table*}

\subsection{Results}
\vspace{-0.3cm}
To reduce the uncertainty of experiments, we run through our experiment codes over five random seeds, including data splitting and few-shots tuning, and report both the mean and standard deviation of each metric on each item in Table~\ref{main_table}. 

\subsubsection{RQ1: MeNSP effectively scores responses based on context and the exemplars without training (zero-shot).} 
Under the zero-shot setting, all the Kappa values of the three items increase from negative to more than 0.30, and the F1 scores also increase among the three items, which indicates the effectiveness of MeNSP in the zero-shot condition (compared with Random baseline).
One of the items (G5) has its Kappa exceed 0.50 with MeNSP, which reaches an acceptable benchmark of the trained model. 
Although MeNSP surpluses a 0.4 Kappa score only on G5, it still achieves at least a 0.3 Kappa score on the other two items, showing the great potential of MeNSP.

\subsubsection{RQ2: MeNSP can be improved by training on a few sample responses on easier items (few-shots).}
To discuss with the few-shots setting, MeNSP performs better than itself on item G4, with the Kappa of random MeNSP increasing from 0.30 (zero-shot) to 0.36 (1-shot), and to 0.38 (3-shots).
However, increasing the number of sample responses does not improve MeNSP further on G5 and G6. Although MeNSP's performance fails on two items, it still has a higher average Kappa and lower standard deviation than all baselines under the 1-shot and 3-shots settings on all three items. This indicates that MeNSP performs better and is more stable than other models. 

Considering the lower complexity level of G4 (level 2) than G5 and G6 (level 3), we first conclude that MeNSP's performance can be improved by training on some responses on the low complexity items. 
However, for items with high complexity, we argue that the performance of MeNSP heavily relies on the characteristics of the sample responses provided for training.
According to our experiments, we observe an improvement of Kappa on G5 from 0.59 to 0.61 with random seed 55301 and from 0.57 to 0.60 with random seed 9, and a gain of Kappa on G6 from 0.34 to 0.35 with random seed 45983. 
A possible speculation of the successful improvement with these random seeds is that the sample responses selected for training are diverse enough\footnote{G5 random samples at grade 2: (1) if the gas bubbles in soda go to the top, then so should the gas particles in the box. (2) Sam could use the Gas bubble idea to support his claim. Gas bubbles are made up of particles. The gas bubbles float to the top of the can. So, the particles in the box will rise to the top also. (3) The gas particles go to the top, like in soda.} to help MeNSP to learn complete and unbiased knowledge about the general scoring rubrics. This speculation also aligns with the finding of RQ3.

\subsubsection{RQ3: The quality of the provided samples affects the few-shots learning performance.} 
We compare MeNSP on item G4 with different sample-gathering strategies (random or manual). The results show that with the 1-shot setting, both strategies lead MeNSP to a similar performance, with the average Kappa close to 0.36. 
However, with the 3-shot setting, the random strategy performs better than the manual strategies, with the average Kappa increase from 0.34 (manual) to 0.38 (random). 
As the random strategy extracts samples from the real student responses, the manual strategy provides augmented student responses similar to the exemplars\footnote{G4 manual samples at grade 1: (1) The evidence suggests that the air in the balloon has spread throughout.
(2) Charlie's claim is right because the air in the balloon spreads everywhere.
(3) That air in the balloon spreads everywhere supports claim.}.
Therefore, combined with the observation of improvements with three random seeds on G5 and G6 discussed earlier, we summarize that the diversity of the given samples is a potential factor that impacts the efficiency of few-shots learning.

\section{Conclusion and Discussion}
In this study, we develop a zero-shot approach (MeNSP) to score student responses automatically. We propose three research questions and examine MeNSP's performance with 1-shot and 3-shot scenarios.
Through experiments, we prove the effectiveness of MeNSP on automatically scoring student responses based on exemplars without training. We also find that increasing training samples can improve MeNSP's performance on items with lower complexity. However, the quality of the improvement is related to the characteristics of the training samples, for example, the diversity of the sample responses. Given that our goal of this study is to preliminary demonstrate the feasibility of MeNSP and the prompt learning methods in automatic scoring science assessment, the machine scoring accuracy may be used for low-stake formative assessment practices. Future research should improve the performance of MeNSP and use varying datasets to verify the approach.

\section*{Acknowledgement}
The study was funded by National Science Foundation(NSF) (Award \# 2101104, 2138854, PI: Zhai). Any opinions, findings, conclusions, or recommendations expressed in this material are those of the author(s) and do not necessarily reflect the views of the NSF.

\bibliographystyle{splncs04}
\bibliography{paper}

\begin{thebibliography}{10}
\providecommand{\url}[1]{\texttt{#1}}
\providecommand{\urlprefix}{URL }
\providecommand{\doi}[1]{https://doi.org/#1}

\bibitem{amerman-a}
Amerman, H., et~al.: Does transformer deep learning yield more accurate sores
  on student written explanations than traditional machine learning? In: AERA

\bibitem{bejar1991methodology}
Bejar, I.I.: A methodology for scoring open-ended architectural design
  problems. Journal of Applied Psychology  (1991)

\bibitem{brown2020language}
Brown, T., Mann, B., Ryder, N., Subbiah, M., Kaplan, J.D., Dhariwal, P.,
  Neelakantan, A., Shyam, P., Sastry, G., Askell, A., et~al.: Language models
  are few-shot learners. Advances in neural information processing systems
  (2020)

\bibitem{national2012framework}
Council, N.R., et~al.: A framework for K-12 science education: Practices,
  crosscutting concepts, and core ideas. National Academies Press (2012)

\bibitem{devlin2018bert}
Devlin, J., Chang, M.W., Lee, K., Toutanova, K.: Bert: Pre-training of deep
  bidirectional transformers for language understanding. In: ACL (2019)

\bibitem{gao2020making}
Gao, T., Fisch, A., Chen, D.: Making pre-trained language models better
  few-shot learners. arXiv preprint arXiv:2012.15723  (2020)

\bibitem{gerard2019guiding}
Gerard, L., Kidron, A., Linn, M.C.: Guiding collaborative revision of science
  explanations. Int. Journal of Computer-Supported Collaborative Learning
  (2019)

\bibitem{harris2019designing}
Harris, C.J., et~al.: Designing knowledge-in-use assessments to promote deeper
  learning. Educational measurement: issues and practice  (2019)

\bibitem{haudek2012they}
Haudek, K.C., et~al.: What are they thinking? automated analysis of student
  writing about acid--base chemistry in introductory biology. Life Sciences
  Education  (2012)

\bibitem{haudek2021exploring}
Haudek, K.C., Zhai, X.: Exploring the effect of assessment construct complexity
  on machine learning scoring of argumentation (2021)

\bibitem{lee2019automated}
Lee, H.S., et~al.: Automated text scoring and real-time adjustable feedback:
  Supporting revision of scientific arguments involving uncertainty. Science
  Education  (2019)

\bibitem{litman2016natural}
Litman, D.: Natural language processing for enhancing teaching and learning.
  In: Thirtieth AAAI conference on artificial intelligence (2016)

\bibitem{liu2014automated}
Liu, O.L., et~al.: Automated scoring of constructed-response science items:
  Prospects and obstacles. Educational Measurement: Issues and Practice  (2014)

\bibitem{liu2021pre}
Liu, P., et~al.: Pre-train, prompt, and predict: A systematic survey of
  prompting methods in natural language processing. arXiv preprint
  arXiv:2107.13586  (2021)

\bibitem{liu2021gpt}
Liu, X., et~al.: Gpt understands, too. arXiv preprint arXiv:2103.10385  (2021)

\bibitem{liu2021p}
Liu, X., et~al.: P-tuning v2: Prompt tuning can be comparable to fine-tuning
  universally across scales and tasks. arXiv preprint arXiv:2110.07602  (2021)

\bibitem{lu2021fantastically}
Lu, Y., et~al.: Fantastically ordered prompts and where to find them:
  Overcoming few-shot prompt order sensitivity. arXiv preprint arXiv:2104.08786
   (2021)

\bibitem{maestrales2021using}
Maestrales, S.e.a.: Using machine learning to score multi-dimensional
  assessments of chemistry and physics. Journal of Science Education and
  Technology  (2021)

\bibitem{mayer2022prompt}
Mayer, C.W., Ludwig, S., Brandt, S.: Prompt text classifications with
  transformer models! an exemplary introduction to prompt-based learning with
  large language models. Journal of Research on Technology in Education  (2022)

\bibitem{nehm2012transforming}
Nehm, R.H., Ha, M., Mayfield, E.: Transforming biology assessment with machine
  learning: automated scoring of written evolutionary explanations. Journal of
  Science Education and Technology  (2012)

\bibitem{omizo2021detecting}
Omizo, R., Meeks, M., Hart-Davidson, W.: Detecting high-quality comments in
  written feedback with a zero shot classifier. In: ACM ICDC (2021)

\bibitem{osborne2016development}
Osborne, J.F., et~al.: The development and validation of a learning progression
  for argumentation in science. Journal of research in science teaching  (2016)

\bibitem{Pellegrino2013-hr}
Pellegrino, J.W.: Proficiency in science: Assessment challenges and
  opportunities. Science  (2013)

\bibitem{powers2015f}
Powers, D.M.: What the f-measure doesn't measure: Features, flaws, fallacies
  and fixes. arXiv preprint arXiv:1503.06410  (2015)

\bibitem{riordan2020empirical}
Riordan, B.e.a.: An empirical investigation of neural methods for content
  scoring of science explanations. In: Proceedings of the fifteenth workshop on
  innovative use of NLP for building educational applications (2020)

\bibitem{salton1988term}
Salton, G., Buckley, C.: Term-weighting approaches in automatic text retrieval.
  Information processing \& management  (1988)

\bibitem{schick2020s}
Schick, T., Sch{\"u}tze, H.: It's not just size that matters: Small language
  models are also few-shot learners. arXiv preprint arXiv:2009.07118  (2020)

\bibitem{schick2020exploiting}
Schick, T., Schütze, H.: Exploiting cloze questions for few-shot text
  classification and natural language inference. Computing Research Repository
  (2020)

\bibitem{shin2020autoprompt}
Shin, T., et~al.: Autoprompt: Eliciting knowledge from language models with
  automatically generated prompts. arXiv preprint arXiv:2010.15980  (2020)

\bibitem{su2022transferability}
Su, Y., et~al.: On transferability of prompt tuning for natural language
  processing. In: NACL. pp. 3949--3969 (2022)

\bibitem{uhl2021introductory}
Uhl, J.D., et~al.: Introductory biology undergraduate students' mixed ideas
  about genetic information flow. Biochemistry and Molecular Biology Education
  (2021)

\bibitem{vu2021spot}
Vu, T., et~al.: Spot: Better frozen model adaptation through soft prompt
  transfer. arXiv preprint arXiv:2110.07904  (2021)

\bibitem{Wolfe2020-ul}
Wolfe, E.W., Wendler, C.L.W.: Why should we care about human raters? Applied
  Measurement in Education  (2020)

\bibitem{wu2023survey}
Wu, X., et~al.: A survey of graph prompting methods: Techniques, applications,
  and challenges. arXiv preprint arXiv:2303.07275  (2023)

\bibitem{zhai2021practices}
Zhai, X.: Practices and theories: How can machine learning assist in innovative
  assessment practices in science education. Journal of Science Education and
  Technology  (2021)

\bibitem{zhai2022assessing}
Zhai, X., Haudek, K.C., Ma, W.: Assessing argumentation using machine learning
  and cognitive diagnostic modeling. Research in Science Education  (2022)

\bibitem{zhai2021validity}
Zhai, X., Krajcik, J., Pellegrino, J.W.: On the validity of machine
  learning-based next generation science assessments: A validity inferential
  network. Journal of Science Education and Technology  (2021)

\bibitem{zhai2021meta}
Zhai, X., Shi, L., Nehm, R.H.: A meta-analysis of machine learning-based
  science assessments: factors impacting machine-human score agreements.
  Journal of Science Education and Technology  (2021)

\bibitem{zhai2020applying}
Zhai, X., Yin, Y., Pellegrino, J.W., Haudek, K.C., Shi, L.: Applying machine
  learning in science assessment: a systematic review. Studies in Science
  Education  (2020)

\bibitem{zhang2022automatic}
Zhang, M., et~al.: Automatic short math answer grading via in-context
  meta-learning. arXiv preprint arXiv:2205.15219  (2022)

\bibitem{zhong2021adapting}
Zhong, R., Lee, K., Zhang, Z., Klein, D.: Adapting language models for
  zero-shot learning by meta-tuning on dataset and prompt collections. In:
  EMNLP (2021)

\end{thebibliography}
\end{document}